\documentclass[conference]{IEEEtran}
\IEEEoverridecommandlockouts
\usepackage{cite}
\usepackage{graphicx}
\usepackage{url}
\usepackage{booktabs,multirow}

\begin{document}

\title{CastClaw: A Human-in-the-Loop Autonomous Agent for Interactive Time Series Forecasting}

\author{\IEEEauthorblockN{Xiaoyu Tao, Mingyue Cheng\textsuperscript{*}\thanks{\textsuperscript{*}Corresponding author: Mingyue Cheng (mycheng@ustc.edu.cn).}, Ze Guo, Bokai Pan, Qi Liu, Shijin Wang, and Enhong Chen}
\IEEEauthorblockA{State Key Laboratory of Cognitive Intelligence, University of Science and Technology of China, Hefei, China\\
txytiny@mail.ustc.edu.cn, \{mycheng, qiliuql, cheneh\}@ustc.edu.cn, sjwang3@iflytek.com}
}

\maketitle

\begin{abstract}
Practical time-series forecasting requires domain knowledge, constraints, plausibility checks, and human intervention. We present \textbf{CastClaw}, an interactive forecasting system centered on an autonomous forecasting loop and supported by a forecasting-oriented harness. Users create and refine tasks in natural language through command-line and Web interfaces. An internal large language model (LLM) agent reads the task and evidence, selects a permitted tool action or request for human feedback, and uses checked results to decide whether to continue, revisit an analysis, switch a candidate, or stop. The harness supplies tool integration, controlled execution, state management, historical validation, and execution limits. npm distribution connects installation and configuration to interactive execution and forecast and report export. Across five public electricity-price datasets, CastClaw achieves the lowest point-estimate MSE and MAE among 16 baselines. A Nord Pool case study illustrates its inspectable decision process, while provincial electricity-load data from North China (January--June 2026) demonstrate offline task execution.
\end{abstract}


\section{Introduction}

Time series forecasting maps historical observations to future values and underpins applications~\cite{box2015time,lago2021forecasting}. Foundation models such as Chronos~\cite{ansari2024chronos}, TimesFM~\cite{das2024decoder}, and Moirai~\cite{woo2024moirai} support pretrained forecasting across datasets. Practical forecasting also requires domain knowledge, constraints, plausibility checks, missing context, and human intervention. Reliable forecasting is therefore a system-level process involving both numerical models and human judgment.

Slow-thinking approaches investigate inference-time reasoning in TimeReasoner~\cite{wang2025timereasoner} and reinforcement fine-tuning in Time-R1~\cite{zhou2025timer1}. Building on reasoning and tool use~\cite{yao2023react,schick2023toolformer}, recent work further explores agentic forecasting with planning, action, reflection, and memory~\cite{cheng2026position}, including experience reuse~\cite{tao2026memcast}, tool-augmented sequential decision making~\cite{tao2026castr1}, and the integration of numerical priors with contextual reasoning and reflective checks~\cite{tao2026castfsr}.
Human--AI collaborative forecasting combines human and machine predictions~\cite{miyoshi2018human}, while AlphaCast~\cite{zhang2025alphacast} supports interactive human--LLM co-reasoning. However, human intervention is not always beneficial: poorly calibrated corrections can reduce forecasting accuracy~\cite{zhang2026calibrated}. These studies motivate a human-in-the-loop forecasting system that couples feedback-driven action selection with domain checks, controlled tool execution, validated revision, and explicit stopping decisions.

We present \textbf{CastClaw}, an interactive, human-in-the-loop \textbf{forecasting system} centered on an autonomous \textbf{forecasting loop}. Users specify targets, horizons, constraints, and hypotheses in natural language through command-line and Web interfaces. An internal LLM agent selects permitted actions from task state and feedback, while a \textbf{harness} executes, constrains, validates, and records them. Historical validation and hard constraints govern candidate adoption, while users can refine or stop execution.
We evaluate CastClaw on five public electricity-price datasets, an inspectable Nord Pool case study, and provincial electricity-load data from North China covering January--June 2026. The live demonstration exposes inputs, tool calls, checks, revisions, and stopping decisions for inspection.

Our contributions are:
(1) \textbf{System:} CastClaw, an interactive forecasting system with npm distribution and command-line and Web interfaces for natural-language task creation, intervention, and result inspection;
(2) \textbf{Engineering:} an autonomous forecasting loop for feedback-driven action selection, supported by a harness for tool execution, state management, human intervention, historical validation, and stopping control; and
(3) \textbf{Demonstration:} quantitative evaluation on five public electricity-price datasets, an inspectable Nord Pool case study, and offline validation on provincial electricity-load data from North China.

\begin{figure*}[!t]
\centering
\includegraphics[width=0.98\textwidth]{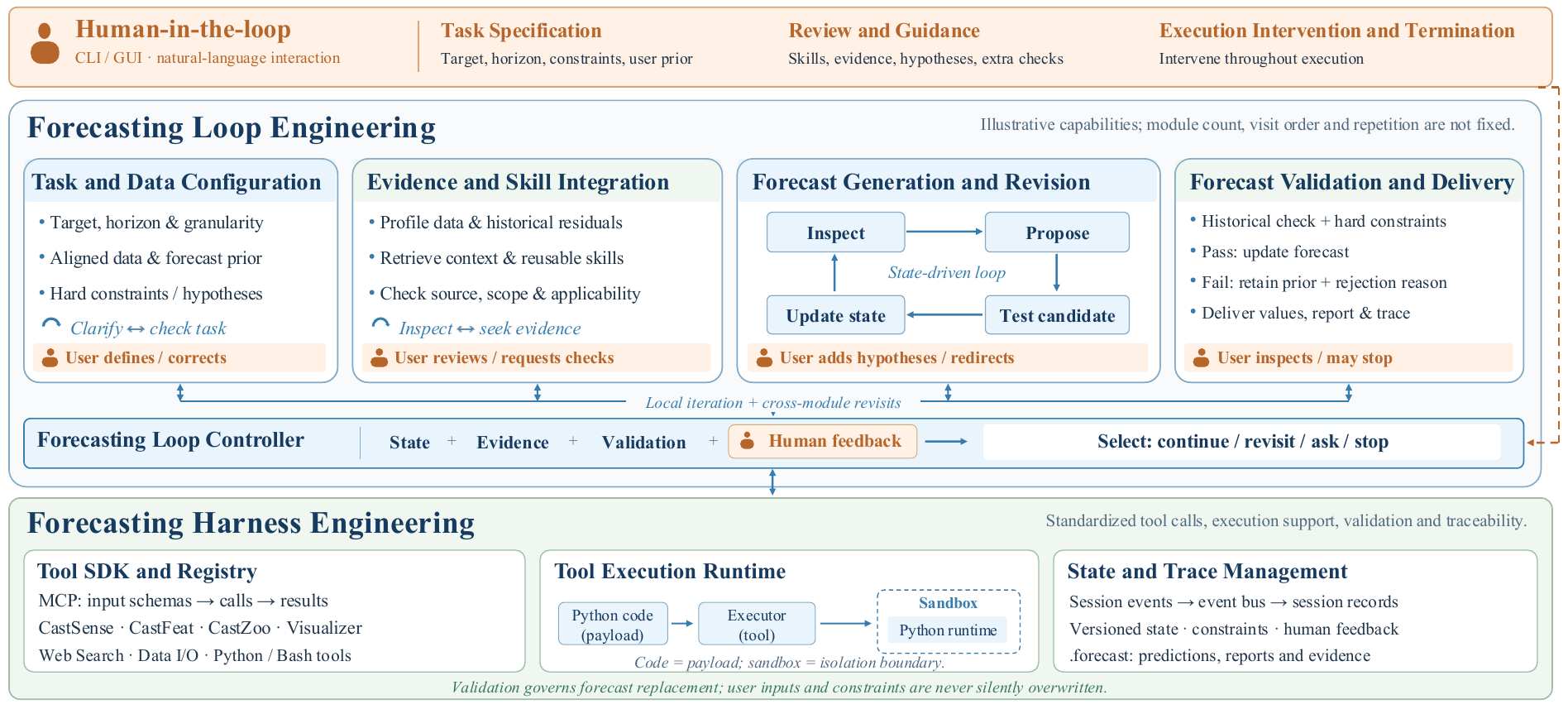}
\caption{CastClaw system architecture: human interaction throughout execution, four illustrative functional groups governed by a forecasting loop controller, and shared harness services for tool calls, execution, state, and trace management.}
\label{fig:castclaw-framework}
\end{figure*}

\section{CastClaw System}

Fig.~\ref{fig:castclaw-framework} organizes CastClaw into three coordinated layers: human interaction, forecasting loop engineering, and shared forecasting harness services.

\subsection{Forecasting Loop Engineering}

The LLM-based Forecasting Loop Controller selects actions based on task state, evidence, validation results, and human feedback. Each iteration reads the current state, selects and executes an action, validates the result, and then continues, revisits, asks, or stops. Missing requirements, data issues, anomalies, model disagreement, or human intervention may trigger further iterations.

The four capability groups, Task and Data Preparation, Evidence Acquisition and Tool Use, Forecast Generation and Revision, and Validation and Result Delivery, do not represent fixed sequential stages. Each group may be revisited over multiple iterations, while their ordering and repetition are dynamically determined by the evolving system state, feedback signals, and phase-gating conditions.

For example, task and data setup checks alignment, missing values, and granularity. Evidence collection profiles residuals and reviews context and applicable Skills. Forecasting and revision iterate \textit{Inspect--Propose--Test candidate--Update state} with a user prior or allowed model. Validation and delivery accept a supported candidate or retain the prior with a rejection reason and report. Failed checks can trigger data inspection, model switching, retrieval, or expert consultation according to the controller's next decision.

\subsection{Harness Engineering for Forecasting Loops}

\noindent\textbf{Tool interfaces and execution.} The registry exposes CastSense, CastFeat, CastZoo, Visualizer, retrieval, and data I/O tools. Model Context Protocol (MCP) interfaces specify input schemas, calls, and results. The runtime view distinguishes Python code as payload, an executor tool, and a Python execution environment containing the interpreter and dependencies; sandbox denotes its depicted isolation boundary.

\noindent\textbf{State and trace management.} The harness records session events, requests, data and model versions, constraints, evidence, candidates, and actions. Corrections mark affected forecasts stale while retaining compatible evidence; user-supplied values and hard constraints cannot be overwritten silently. Skills and past runs retain sources and applicability conditions for review before reuse; conflicts with current data or constraints lead to rejection.

\noindent\textbf{Human intervention.} The harness records task corrections, evidence reviews, and execution interventions with their source, time, and scope for the next loop decision. Hypotheses remain testable, and proposed adjustments require validation.

\noindent\textbf{Historical validation.} Candidate adjustment rules are tested on historical validation windows before application to the current forecast; future targets are unavailable at decision time. A rule is adopted only if it passes the validation criterion and the resulting forecast satisfies hard constraints. Otherwise, the active forecast and the rejected candidate's record are retained.

\noindent\textbf{Limits and reporting.} Phase gates and the trial cap (\texttt{max\_experiments}) bound exploration. Repeated trials without new experience pause for human review at \texttt{no\_improve\_threshold}. Expert queries also pause for feedback; a supported forecast, budget exhaustion, or user stop ends the run. Reports preserve decisions, warnings, tested forecasts, interventions, and evidence with provenance labels.

\begin{table*}[t]
	\centering
	\caption{Test-set forecasting performance on five electricity-price datasets (BE, DE, FR, NP, and PJM). Lower is better. Best results are in \textbf{bold}; second-best results are \underline{underlined}.}
	\label{tab:forecasting_results}
	\scriptsize
	
		\textit{(a) Baselines (part I)}\par\vspace{1pt}
	\resizebox{\textwidth}{!}{%
		\begin{tabular}{@{}l|cc|cc|cc|cc|cc|cc|cc|cc@{}}
			\toprule
			\multirow{2}{*}{Dataset}
			& \multicolumn{2}{c|}{ARIMA} & \multicolumn{2}{c|}{Prophet}
			& \multicolumn{2}{c|}{DLinear} & \multicolumn{2}{c|}{ConvTimeNet}
			& \multicolumn{2}{c|}{PatchTST} & \multicolumn{2}{c|}{iTransformer}
			& \multicolumn{2}{c|}{TimeXer} & \multicolumn{2}{c}{TimesFM} \\
			& MSE & MAE & MSE & MAE & MSE & MAE & MSE & MAE
			& MSE & MAE & MSE & MAE & MSE & MAE & MSE & MAE \\
			\midrule
			BE
			& 1.978 & 0.814 & 2.321 & 0.918 & 1.697 & 0.677 & 1.482 & 0.581
			& 1.905 & 0.647 & 1.684 & 0.569 & 1.437 & 0.502 & 1.349 & 0.510 \\
			DE
			& 3.269 & 1.076 & 2.103 & 0.986 & 1.000 & 0.699 & \underline{0.964} & 0.665
			& 0.983 & 0.666 & 1.298 & 0.697 & 1.009 & 0.708 & 0.967 & \underline{0.660} \\
			FR
			& 6.410 & 1.189 & 6.831 & 1.280 & 6.369 & 1.000 & 5.376 & 0.800
			& 5.021 & 0.740 & 6.066 & 0.869 & 5.578 & 0.827 & \underline{4.580} & 0.651 \\
			NP
			& 1.798 & 0.801 & 0.707 & 0.583 & 0.431 & 0.446 & 0.393 & 0.411
			& 0.454 & 0.441 & 0.373 & 0.404 & 0.412 & 0.416 & 0.317 & \underline{0.361} \\
			PJM
			& 0.907 & 0.710 & 0.410 & 0.500 & 0.286 & 0.387 & 0.226 & 0.342
			& 0.190 & 0.323 & 0.238 & 0.344 & 0.197 & 0.318 & 0.207 & 0.324 \\
			\bottomrule
		\end{tabular}%
	}
	
	\vspace{5pt}
		\textit{(b) Baselines (part II) and CastClaw}\par\vspace{1pt}
	\resizebox{\textwidth}{!}{%
		\begin{tabular}{@{}l|cc|cc|cc|cc|cc|cc|cc|cc|cc@{}}
			\toprule
			\multirow{2}{*}{Dataset}
			& \multicolumn{2}{c|}{Sundial} & \multicolumn{2}{c|}{Time-LLM}
			& \multicolumn{2}{c|}{TokenCast} & \multicolumn{2}{c|}{S$^2$IP-LLM}
			& \multicolumn{2}{c|}{PromptCast} & \multicolumn{2}{c|}{TimeReasoner}
			& \multicolumn{2}{c|}{TimeSeriesScientist} & \multicolumn{2}{c|}{AlphaCast}
			& \multicolumn{2}{c}{CastClaw} \\
			& MSE & MAE & MSE & MAE & MSE & MAE & MSE & MAE & MSE & MAE
			& MSE & MAE & MSE & MAE & MSE & MAE & MSE & MAE \\
			\midrule
			BE
			& 1.478 & 0.513 & 1.758 & 0.701 & 1.485 & 0.547 & 1.591 & 0.567 & 1.870 & 0.637
			& \underline{1.299} & \underline{0.466} & 2.057 & 0.712 & 1.484 & 0.510 & \textbf{1.228} & \textbf{0.401} \\
			DE
			& 1.155 & 0.700 & 1.025 & 0.707 & 0.973 & 0.702 & 1.315 & 0.722 & 1.229 & 0.708
			& 1.082 & 0.760 & 1.028 & 0.673 & 1.043 & 0.662 & \textbf{0.674} & \textbf{0.522} \\
			FR
			& 6.105 & 0.682 & 6.450 & 0.971 & 6.363 & 0.961 & 5.765 & 0.888 & 7.302 & 1.173
			& 4.644 & \underline{0.650} & 5.977 & 1.033 & 5.318 & 0.740 & \textbf{4.058} & \textbf{0.497} \\
			NP
			& 0.362 & 0.378 & 0.422 & 0.447 & 0.412 & 0.411 & 0.389 & 0.391 & 0.565 & 0.449
			& 0.368 & 0.386 & 0.606 & 0.476 & \underline{0.316} & 0.365 & \textbf{0.260} & \textbf{0.276} \\
			PJM
			& 0.201 & 0.324 & 0.253 & 0.368 & 0.215 & 0.319 & 0.237 & 0.339 & 0.316 & 0.454
			& 0.207 & 0.328 & 0.302 & 0.439 & \underline{0.176} & \underline{0.292} & \textbf{0.159} & \textbf{0.259} \\
			\bottomrule
		\end{tabular}%
	}
\end{table*}

\section{Interactive demonstration}

\subsection{System Interfaces and Deployment}

CastClaw is distributed through npm. Run \texttt{npm install -g castclaw}, configure an LLM provider, and prepare the dependencies required by the selected forecasting tools. Launch \texttt{castclaw} for the terminal interface or \texttt{castclaw web} for the Web interface, then describe the dataset, target, horizon, and constraints in natural language. Prerequisites, configuration instructions, sample tasks, and a demonstration video are available at \url{https://ustc-time-series.github.io/cast-claw/}.
Users inspect evidence, tool outputs, and loop decisions and can intervene before obtaining the result. Forecast artifacts, evaluation metrics, and the execution report are saved under \texttt{.forecast/} for inspection and export. MCP interfaces connect additional tools to the same execution.

\subsection{Human interaction during a run}

Through either interface, users can correct the target or horizon, add a constraint or hypothesis, request another check, or stop the run. They can inspect the resulting evidence, warnings, unresolved questions, and forecast changes in the report. A representative human--agent run is shown in Fig.~\ref{fig:castclaw-ui}.

\section{Evaluation and case study}

\subsection{Experimental settings}

We evaluate CastClaw on five public electricity-price datasets: BE, DE, FR, NP, and PJM~\cite{lago2021forecasting}. Each uses a chronological 7:1:2 train/validation/test split. The 16 baselines in Table~\ref{tab:forecasting_results} cover statistical methods~\cite{box2015time,taylor2018forecasting}, neural models such as ConvTimeNet~\cite{cheng2024convtimenet}, foundation models~\cite{das2024decoder,liu2025sundial}, LLM forecasters~\cite{jin2024timellm,xue2022promptcast,tao2025tokencast}, reasoning-based forecasting~\cite{wang2025timereasoner}, and forecasting agents such as AlphaCast~\cite{zhang2025alphacast}. We report test MSE and MAE for the complete CastClaw system. Test targets are never used for selection or revision, context is restricted to pre-origin information, and the candidate pool is fixed using validation data before test evaluation.

\subsection{Main table analysis}

CastClaw achieves the lowest MSE and MAE across all ten dataset--metric cells in Table~\ref{tab:forecasting_results}, with average reductions of 14.9\% in MSE and 18.6\% in MAE over the strongest competing results. These are point estimates for the complete system after validation-based candidate checking. Repeated controlled runs are needed to assess statistical significance and attribute gains to individual components, including tool calls, human interventions, and stopping rules.

\subsection{Representative Nord Pool run}

\begin{figure*}[!t]
\centering
\includegraphics[width=0.9\textwidth]{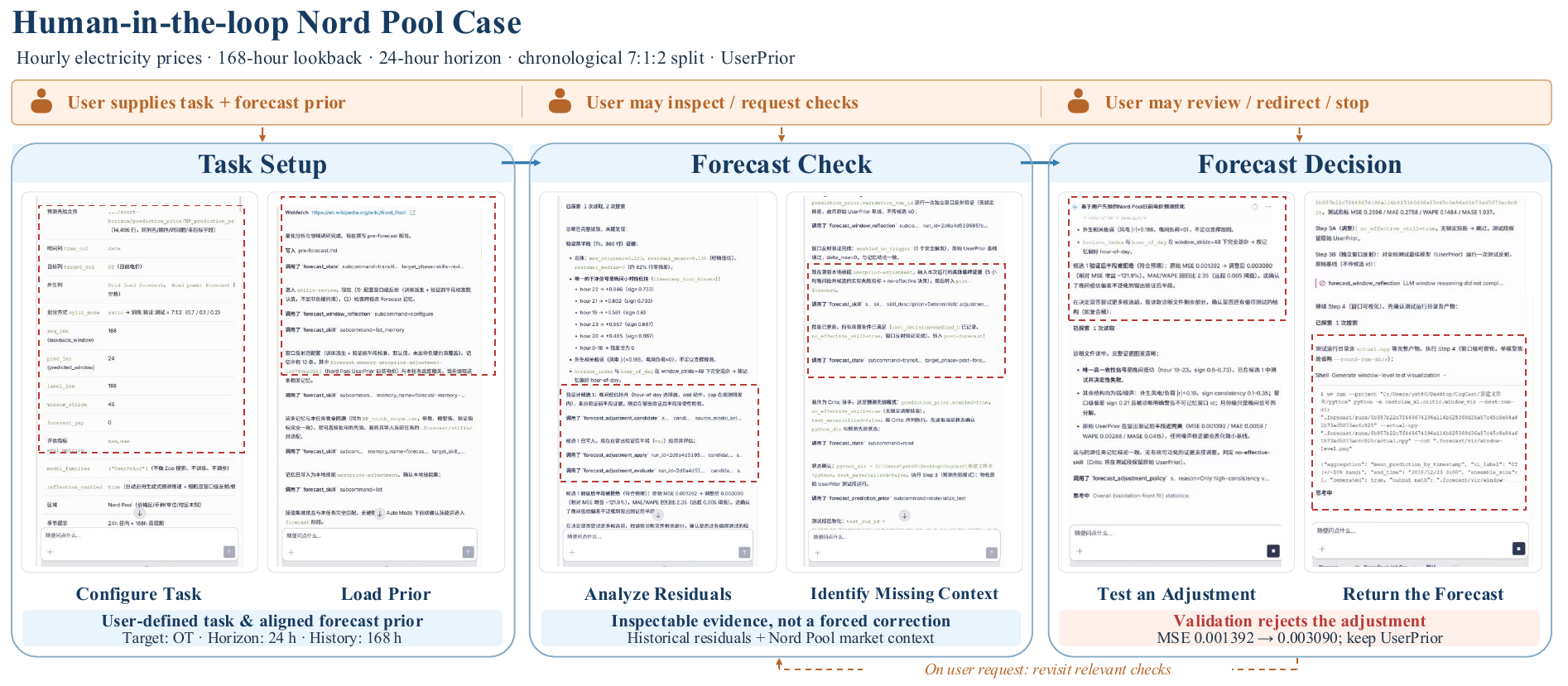}
\caption{Nord Pool case: Task Setup records UserPrior (left); Forecast Check examines residuals and context (center); Forecast Decision rejects an adjustment and retains UserPrior (right). The dashed path denotes a recheck on user request.}
\label{fig:castclaw-ui}
\end{figure*}

Fig.~\ref{fig:castclaw-ui} shows task setup, forecast checking, and decision in a Nord Pool run with a 168-hour history, a 24-hour horizon, and a chronological 7:1:2 split. The user supplies an aligned forecast prior. CastClaw records the task, checks historical residuals, retrieves market context, and constructs a candidate adjustment rule.
The displayed adjustment increases validation MSE from 0.001392 to 0.003090; it is rejected, and UserPrior is retained. The decision panel records compared versions, the rejection reason, stopping decision, and artifacts.

\subsection{Offline execution validation}

CastClaw was also validated offline on provincial electricity-load data from North China (January--June 2026) using the same harness with load-specific resources. This validates offline task execution only, not comparative accuracy or production deployment.

\noindent\textit{Scope, ethics, and data use.} The aggregate load data contain no human-subject information and were used under provider authorization and confidentiality requirements. They are excluded from Table~\ref{tab:forecasting_results} and support no claims of business impact.

\section{Conclusion}

CastClaw is an interactive forecasting system combining an autonomous forecasting loop, a forecasting-oriented harness, and npm-distributed command-line and Web interfaces. The internal agent selects permitted actions from feedback; the harness executes, constrains, validates, and records them. Public electricity-price experiments show the lowest point-estimate MSE and MAE among the evaluated methods; the Nord Pool case and provincial-load validation illustrate inspectable decision-making and offline task execution. Ablations and user studies remain future work.

\section*{Acknowledgment}
This work was supported by grants from the National Natural Science Foundation of China (62502486), the Fundamental Research Funds for the Central Universities of China, and the USTC Kunpeng-Ascend Scientific and Education Innovation Excellence Center.

\bibliographystyle{IEEEtran}
\bibliography{references}

\end{document}